\documentclass[finalnew]{agujournal2019}
\usepackage{url} 
\usepackage[finalnew]{trackchanges}
\usepackage{soul}
\soulregister\cite7 
\soulregister\citeA7 
\soulregister\ref7

\usepackage{amsmath}

\draftfalse

\journalname{JGR: Machine Learning and Computation}

\begin{document}


\title{Neural Surrogate HMC: On Using Neural Likelihoods for Hamiltonian Monte Carlo in Simulation-Based Inference}


\def\authorEmail{linneamw@hawaii.edu}

\authors{L. M. Wolniewicz\affil{1}, P. Sadowski\affil{1}, C. Corti\affil{2,3,4}}

\affiliation{1}{Information and Computer Sciences, University of Hawai`i at Mānoa, 1680 East-West Road, Honolulu HI, USA}
\affiliation{2}{Physics and Astronomy, University of Hawai`i at Mānoa, 2505 Correa Road, Honolulu HI, USA}
\affiliation{3}{NASA Goddard Space Flight Center, 8800 Greenbelt Road, Greenbelt MD, USA}
\affiliation{4}{Universities Space Research Association, 425 3rd Street SW, Suite 950, Washington, DC, USA}

\correspondingauthor{Linnea M. Wolniewicz}{linnea@wolniewicz.com}


\begin{keypoints}
\item Neural surrogate HMC provides fast, stable, and differentiable likelihood evaluations to enable simulation-based inference.
\item We give practical guidance on when to use neural surrogate HMC, and how to assess the convergence of the neural likelihood model. 
\item We demonstrate our method by achieving state-of-the-art constraints on the heliospheric transport of galactic cosmic rays.

\vspace{\baselineskip}
\vspace{\baselineskip}
\vspace{\baselineskip}
\vspace{\baselineskip}
\vspace{\baselineskip}
\vspace{\baselineskip}
\vspace{\baselineskip}
\vspace{\baselineskip}
\vspace{\baselineskip}
\vspace{\baselineskip}
\vspace{\baselineskip}
\vspace{\baselineskip}
\vspace{\baselineskip}
\vspace{\baselineskip}
\vspace{\baselineskip}
\vspace{\baselineskip}
\vspace{\baselineskip}
\vspace{\baselineskip}
\vspace{\baselineskip}
\vspace{\baselineskip}
\vspace{\baselineskip}
\vspace{\baselineskip}
\vspace{\baselineskip}
\vspace{\baselineskip}
\vspace{\baselineskip}
\vspace{\baselineskip}
Accepted for publication in the Journal of Geophysical Research: Machine Learning and Computation. Published 2026 American Geophysical Union. Further reproduction or electronic distribution is not permitted.
\end{keypoints}


\begin{abstract}
Bayesian inference methods such as Markov Chain Monte Carlo (MCMC) typically require repeated computations of the likelihood function, but in some scenarios this is infeasible and alternative methods are needed.
Simulation-based inference (SBI) methods address this problem by using machine learning to amortize computations. In this work, we highlight a particular synergy between the SBI method of neural likelihood estimation and the classic MCMC method of Hamiltonian Monte Carlo.
We show that approximating the likelihood function with a neural network model can provide three distinct advantages: (1) amortizing the computations for MCMC; (2) providing gradients for Hamiltonian Monte Carlo, and (3) smoothing over noisy simulations resulting from numerical instabilities. We provide practical guidelines for defining a prior, sampling a training set, and evaluating convergence. The method is demonstrated in an application modeling the heliospheric transport of galactic cosmic rays, where it enables efficient inference of latent parameters in the Parker equation.  
\end{abstract}

\section*{Plain Language Summary}
Many problems in the physical sciences can be framed as ``inverse problems'', in which the goal is to predict some hidden variables from observations produced by a physical system that can be modeled by a simulator. If the simulator is stochastic or computationally expensive, solving the inverse problem requires approximations. In this work, we describe why a particular approach --- approximating the simulator with a neural network and using Hamiltonian Monte Carlo to obtain predictions of the hidden variables -- is especially effective under certain conditions. We identify these conditions, provide practical recommendations for its use and evaluation of the result, and demonstrate the method by obtaining state-of-the-art constraints on the hidden parameters of the Parker equation for modeling space weather. Our results and practical recommendations for avoiding common pitfalls will be of interest to those working on inverse problems in the physical sciences.


\section{Introduction}

In the physical sciences, traditional Bayesian inference of latent model parameters is often impractical due to limitations of the physical model. Computing the likelihood may be very expensive or infeasible, as is the case when one can only draw samples from the likelihood using a stochastic, black-box simulator. In such scenarios, machine learning is increasingly used to approximate Bayesian inference in a collection of methods known as simulation-based inference (SBI). SBI uses machine learning to interpolate in the observation and/or parameter space and amortize computations to make inference more scalable. The choice of SBI method depends heavily on the details of the application~\cite{deistler2025simulation}, such as the properties of the simulator and the dimensionality of the parameter and observation spaces.

In this work, we focus on a particular combination of SBI and Bayesian inference methods --- neural likelihood estimation (NLE) combined with Hamiltonian Monte Carlo (HMC) --- that synergize well in certain scenarios. We characterize the conditions under which this would be preferred over other methods and provide practical guidelines for its use. In particular, we highlight the advantage of this approach in applications where the simulator requires numerically solving a partial differential equation, resulting in simulations that are expensive, unstable, and/or black-box (non-differentiable). Our proposed approach is a neural likelihood estimator $q_\phi(x|\theta)$ (a neural network, NN, with parameters $\phi$) trained on data generated by the simulator, $\{(\theta_n, x_n)\}_{n \in 1 \ldots M}$, where each of $M$ examples consists of a parameter $\theta$ and corresponding simulator-generated observation $x$. We then use $q_\phi(x|\theta)$ as a replacement for the true likelihood function to accelerate HMC --- taking advantage of the fact that this surrogate likelihood is fast, numerically-stable, and differentiable. This approach has practical advantages over alternative SBI or even likelihood-based methods under certain conditions.


In experiments, we demonstrate our approach by applying it to the problem of modeling the heliospheric transport of galactic cosmic rays (GCRs)~\cite{bib:Rankin-etal-2022, bib:Engelbrecht-etal-2022}, a component of space weather that will influence the scheduling of future crewed space missions. This application has all three properties that make neural surrogate HMC appealing: the simulator is a slow numerical solver which is non-differentiable and fails a fraction of the time due to numerical instabilities. Posterior inference is performed using HMC with all likelihood evaluations performed by the NLE model --- which has an inductive bias towards smoothness --- trained on a large dataset of precomputed simulator runs. We show that this approach provides significant improvement in accuracy for a fixed computational budget compared to other methods.  Using this approach, we obtain state-of-the-art constraints on the global heliospheric transport parameters.

The contributions of this work are thus:
\begin{enumerate}
    \item Identifying three distinct advantages of neural surrogate HMC for SBI.
    \item Identifying conditions under which neural surrogate HMC provides advantages over other inference methods.
    \item Specific recommendations for defining a prior and training data set.
    \item Specific recommendations for handling the approximation error of the surrogate likelihood.
    \item Demonstrating the method by obtaining state-of-the-art constraints on parameters for the Parker equation for modeling GCR transport in the heliosphere.
\end{enumerate}

\section{Related Work}

Previous work suggesting the use of surrogate models to accelerate HMC goes back at least to~\citeA{rasmussen2003gaussian}, who proposed using a Gaussian process (GP) to model the log posterior. \citeA{zhang2017hamiltonian} proposed to do the same with an extreme learning machine. Other works have used NNs to approximate the gradient of the likelihood~\cite{li2019neural}, predict the HMC transition kernel~\cite{levy2018}, and provide a trainable kernel within the dynamic update of the HMC~\cite{foreman2021flows}. In all these methods, the true likelihood is still used when deciding whether to accept a sample, which ensures samples from the true posterior but requires access to reliable likelihood calculations during inference.

The scenario in which the likelihood is intractable, such as when the simulator can only generate a sample from the conditional $p(x|\theta)$, is known as ``likelihood-free'' inference, or SBI. One SBI method is to train an NLE and use it as a drop-in replacement for the likelihood in MCMC~\cite{deistler2025simulation}. While this discards the theoretical guarantees for the MCMC samples, it enables their use in scenarios where the experimenter would otherwise be forced to resort to approximate Bayesian computation~\cite{diggle1984monte}, which doesn't scale well to higher-dimensional parameters and observations. Even in situations where the true likelihood function can be computed, the NLE+MCMC approach may be preferable to enable portable and fast inference software packages. The first paper to propose HMC using an NLE (i.e. a neural surrogate) for SBI is~\citeA{papamakarios2019sequential}, which suggests a sequential algorithm in which the NLE is repeatedly retrained with additional samples from a refined posterior in order to minimize calls to the simulator. However, the sequential approach proposed in that paper is not necessary in applications where posteriors must be obtained for many observations, which justifies computing a large training set of simulations up front and training a single NLE, further amortizing the cost of inference.

\section{Methods}

This section describes the proposed approach. We begin with a formal specification of the problem, describe the conditions that should be met when using this approach, and provide practical guidance for generating a training set and evaluating the results. 

\subsection{Problem Specification}

\begin{figure*}[h!]
   \centering
   \makebox[1\textwidth][c]{\includegraphics[width=1.1\textwidth]{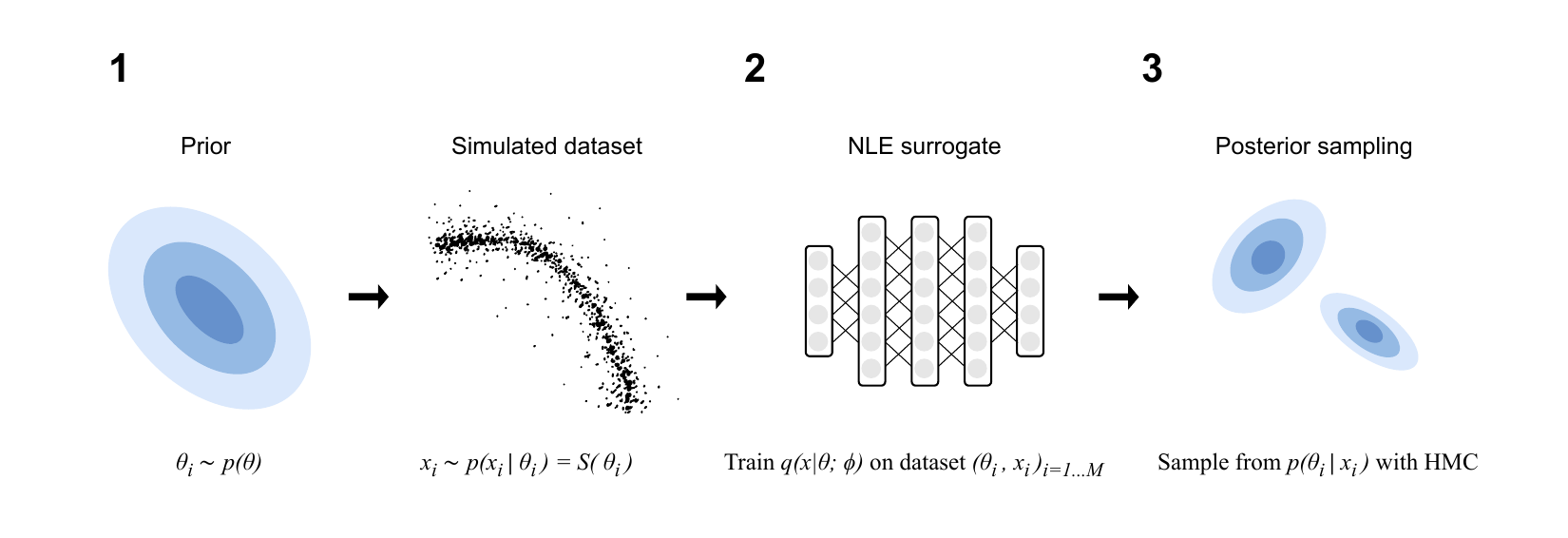}}
   \caption{Our three-step approach to sampling from the posterior $p(\theta|x)$: (1) produce a dataset $(\theta_i, x_i)_{i=1 \ldots M}$ by drawing parameters from the prior $p(\theta)$ and running simulator $\mathcal{S}$ forward. (2) Train an NLE $q(x | \theta ; \phi)$ on the simulated dataset to approximate the simulator. (3) Sample from the posterior $p(\theta_i | x_i)$ for new observations $x_i$ with HMC, using the NLE for all likelihood evaluations. Adapted from \citeA{deistler2025simulation}.
   }
   \label{fig:diagram}
\end{figure*}

We consider a set of $N$ SBI queries for the same simulator. We are given a set of $N$ observations $\{x_1, \ldots, x_N\}$, a prior distribution on hidden parameter(s) $\theta$, and a simulator $\mathcal{S}$ that relates random variables $x$ and $\theta$ by providing samples from a conditional density $p(x | \theta)$ as a function of $\theta$. Both observation $x$ and $\theta$ can be multidimensional. The simulator allows us to sample from the joint distribution $p(x, \theta)$ by drawing parameters from the prior distribution $p(\theta)$ and running $\mathcal{S}$ forward:

\begin{align}
\theta_i &\sim p(\theta) \\
x_i = \mathcal{S}(\theta_i) &\sim p(x_i | \theta_i)
\end{align}

\noindent
The task is to efficiently obtain the posterior $p(\theta_n | x_n)$ for each observation $x_n$ where $n\in 1 \ldots N$. We propose a three step approach (Figure~\ref{fig:diagram}) in which (1) the simulator is used to produce a data set $(\theta_i, x_i)_{i=1 \ldots M}$; (2) a machine learning model $q(x | \theta ; \phi)$, parameterized by $\phi$, is trained to approximate the simulator (an NLE); and (3) the NLE is used to perform HMC, resulting in samples from the posterior.


\subsection{When to use Neural Surrogate HMC}

We argue that this approach is preferable to other inference methods when the problem meets the following conditions:

\begin{enumerate}
    \item \textbf{When computing the likelihood inside the HMC loop is a problem.} The likelihood is sometimes intractable, or its use in the HMC might be undesirable for other reasons. For example, numerical instabilities that lead to errors might make the likelihood computation unreliable, or there might be practical difficulties in including the simulator in a software package for performing the inference. The trade-off here is that one loses the theoretical guarantees that the samples come from the true distribution, so the user should be prepared to evaluate the posterior samples empirically. We provide guidance for such empirical evaluation.
    \item \textbf{When $N$ is large.} Large $N$ justifies pre-computing a large training dataset for the NLE upfront, as opposed to the sequential approach proposed in \citeA{papamakarios2019sequential}. This trade-off will be application-specific, since the size of the training set needed to achieve a particular level of accuracy will depend on the smoothness of the likelihood function.
    \item \textbf{When approximating the likelihood is simpler than approximating the posterior.} Neural posterior estimation (NPE) is another SBI method that directly estimates the posterior $p(\theta|x)$ using a neural network. It does not require MCMC, so it is much faster per inference, but it may require a more complex NN to model the joint distribution over a (potentially) high-dimensional parameter space (e.g. with normalizing flows~\cite{papamakarios2021normalizing}). 
    The decision to use NLE vs. NPE will typically depend on whether it is easier to learn the likelihood or the posterior, and it will depend on the dimensionality and other factors. In general, it is easier to learn a mapping from high dimensions (such as an image) to a lower-dimensional space (such as a class label). There may also be application-specific knowledge that influences this choice. One common reason to prefer the NLE is when the form of the aleatoric uncertainty is understood and can be modeled well (for example, when the observation error is known to be Gaussian, even if the variance is unknown and must be learned from data). 
    \item \textbf{When a neural network is preferable to a Gaussian process (GP).} \citeA{rasmussen2003gaussian} proposed approximating the posterior using a GP. This may be preferable when the inductive biases of a GP are more appropriate for the data or when active learning is used to reduce the number of simulations needed for training. However, neural networks scale \remove{better }to large training sets better than GPs (such as the dataset of 2M simulations demonstrated in this work), and the speedups could be substantial when used in the inner loop of HMC sampling.
\end{enumerate}

\subsection{Generating a Training Set for the NLE}

Generating a training set does not need to match the prior for any particular observation. (This is in contrast to NPE, where the model incorporates information about the prior.) However, we expect the accuracy of the NN model to degrade in regions of parameter space far from the training data. Therefore, the training set should cover the full space of parameter values that we will consider during inference. 

\subsection{Specifying the Prior for HMC}

HMC attempts to efficiently explore the latent parameter space. However, because our surrogate likelihood model has limited ability to generalize, we can use the prior to help constrain the HMC sampler to only explore a \textit{trusted region} of parameter space for which the NLE model should be considered reliable. Thus, the prior probability should rapidly approach or equal zero for parameters outside the training set.

\subsection{Evaluating the Results}
\label{sec:evaluating-the-results}

The theoretical guarantees of MCMC are lost because our method relies on an approximation to the likelihood. However, we can use empirical methods to evaluate our method for a given application. We suggest performing two types of evaluations, one to ensure that the posterior is accurate and one to check that the approach has converged to a stable solution. The former can be achieved with a posterior predictive test, where held-out simulations are used to ensure that the predicted posterior includes the true parameter value. The latter can be achieved by repeating the experiment for a given observation to ensure that it results in the same posterior. For this, we suggest bootstrap sampling from the training dataset, training the NN with different initializations, and HMC sampling with different seeds. The resulting HMC samples can then be compared using the Gelman-Rubin statistic, which is commonly used for testing for MCMC convergence~\cite{Gelman2013Bayesian}. In experiments, we demonstrate the importance of bootstrap sampling and different NN initializations for assessing convergence (Section~\ref{sec:assess_convergence}).




\section{Application: Constraining the Global Heliospheric Transport of GCRs}

We demonstrate our method by using it to constrain five parameters characterizing the transport of GCRs within the heliosphere based on observations by detectors in low Earth orbit. GCRs constitute a major radiation hazard for deep-space human exploration, and understanding their behavior will be critical for crewed missions to the Moon, Mars, and beyond.

\subsection{Description of the Problem}
\label{sec:description_of_problem}

The transport of GCRs within the heliosphere is described by the Parker equation~\cite{bib:parker65:modulation}:

\begin{equation} \label{eqn:parker}
   \frac{\partial J}{\partial t} = -\mathbf{V}_{sw} \cdot \boldsymbol{\nabla}J + \boldsymbol{\nabla} \cdot \left( \boldsymbol{\mathsf{K}} \boldsymbol{\nabla}J \right) + \frac{\boldsymbol{\nabla} \cdot \mathbf{V}_{sw}}{3}  \beta R^{3} \frac{\partial}{\partial R} \left( \frac{J}{\beta R^{2}} \right)
\end{equation}

\noindent
where $J(\mathbf{r}, R)$ is the measured GCR flux at a given position, $\mathbf{r}$, and rigidity, $R$ = (particle momentum) / (particle charge), and $\beta$ is the particle speed divided by the speed of light.
The first term on the right-hand side represents the advection of GCRs by the solar wind with speed $\mathbf{V}_{sw}$.
The second term characterizes the interaction of GCRs with the heliospheric magnetic field (HMF).
$\boldsymbol{\mathsf{K}}$ is the diffusion tensor, which describes the GCR scattering and drifting due to HMF small- and large-scale structures.
It can be separated into a symmetric part, related to diffusion processes parallel and perpendicular to the HMF, and an asymmetric part, related to drift motions along the HMF curvature, gradients, and the wavy heliospheric current sheet (HCS).
The last term describes the adiabatic energy losses (gains) due to a positive (negative) divergence of the solar wind speed, corresponding to the solar wind expansion (compression).

The Parker equation is not analytically solvable, except under specific one-dimensional approximations \cite{bib:Moraal-2013}, therefore numerical modeling approaches have always been the primary way to investigate the transport of GCRs in the heliosphere.
While most of the coefficients of the equation can be derived from observations (e.g., solar wind speed and HMF) or from first principles (e.g., analytical expressions of the diffusion tensor from turbulence theory), there remains some numerical parameters to be constrained from GCR observations, including the GCR flux at the outer heliosphere boundary and the time variation of the diffusion tensor.
Depending on the assumptions and used parametrizations, the number of these parameters can be large enough to make standard regression approaches intractable.
Historically, this issue has been addressed by trying to reduce the dimensionality and manually tuning the remaining free parameters to match the observations.
In the last decade, with the advent of more accurate GCR data, some efforts to systematize the parameter tuning have been made, e.g., least-square minimization using a coarse grid-based search \cite{bib:corti19:numericalmodeling} and MCMC inference with Metropolis-Hastings sampling \cite{bib:Song-etal-2021}.

Our numerical model (simulator $\mathcal{S}$) solves the Parker equation for a given set of parameters, predicting the GCR flux at different rigidities at all locations in the heliosphere on a given spatial grid defined in spherical coordinates.
This is the same quantity measured by GCR instruments, so it can be directly compared to experimental observations.
In particular, this means that we can use the same forward model simulation to generate a training set for any arbitrary GCR instrument.
The coefficients in the Parker equation are parametrized as in \citeA{bib:corti19:numericalmodeling}.
In particular, the solar wind speed, $\mathbf{V}_{sw}$, is radially constant, with an acceleration to supersonic speed close to the Sun, a drop to subsonic speed at the termination shock, and a latitudinal transition to faster speed at the poles. The HMF is a Parker field, as modified in \citeA{bib:Smith-Bieber-1991}, while the HCS is modeled as in \citeA{bib:Kota-Jokipii-1983}. The HMF is characterized by the intensity near Earth, $I_{HMF}$, the solar dipole tilt angle, $\alpha$, and the direction at the solar North pole (positive or negative polarity).
The diffusion coefficients are inversely proportional to the HMF and follow a double power-law in rigidity, separately for the directions parallel and perpendicular to the HMF.
This parametrization is characterized by a normalization constant, $k^{0}_{\parallel}$, and by the rigidity slopes of the parallel and perpendicular diffusion coefficients at low ($a_{\parallel}, a_{\perp}$) and high ($b_{\parallel}, b_{\perp}$) rigidities.

Observations come from the PAMELA and AMS-02 instruments (PAMELA monthly protons are available at the ASI Cosmic Ray Database: \url{https://tools.ssdc.asi.it/CosmicRays/} and AMS-02 daily protons are available at: \url{https://ams02.space/sites/default/files/publication/202105/table-s1-s2824.csv}). PAMELA measured H fluxes at a monthly cadence between 0.4-50 GV in the period 2006-2014 \cite{bib:adriani13:pamela-p-monthly-flux,bib:martucci18:pamela-p-monthly-flux}, while AMS-02 measured H fluxes at daily cadence between 1-100 GV in the period 2011-2019 \cite{bib:aguilar21:periodicities-daily-proton}.
%
Observational noise is assumed to be normally distributed with a standard deviation equal to the sum of the total statistical and systematic uncertainties, common in cosmic ray experimental physics and generally valid for most systematic sources. Although the number of observed cosmic rays is Poisson distributed, the large number of observed events justifies the use of the Gaussian approximation for the statistical uncertainties. Assuming independence of the uncertainties, this gives an exact likelihood obtained by summing the difference between the GCR fluxes predicted by the model \change{($\hat{x}=q(\theta; \phi)$)}{($S(\theta))$} and the observed GCR flux ($x$) scaled by uncertainty $\sigma$ for each observed rigidity bin ($1 \ldots R$):

\begin{align} \label{eq:likelihood}
p(x|\theta) &= \prod_{r \in 1 \ldots R} \frac{1}{\sqrt{2 \pi} \sigma_{r}} \exp\left(-\frac{(x_r - S(\theta)_r)^2}{2\sigma_r^2}\right)\\
-\log p(x|\theta) &\propto \sum_{r \in 1 \ldots R} \frac{(x_r - S(\theta)_r)^2}{\sigma_r^2}
\end{align}

Unfortunately, the simulator is slow, non-differentiable, and suffers from numerical instabilities which sometimes lead to extreme flux values. The simulations with instabilities can usually be detected and removed, but the simulator likelihood should not be trusted absolutely, and an NLE with an inductive bias towards smoothness could actually be more accurate. We also wish to perform inference over many time intervals ($N=133$), so we prefer to simulate training data once, then have a relatively fast inference procedure. This motivates the use of neural surrogate HMC.



\subsection{NLE Setup}
\label{sec:nle_setup}

\begin{figure}[h!]
   \makebox[1\textwidth][c]{\includegraphics[width=1.1\textwidth]{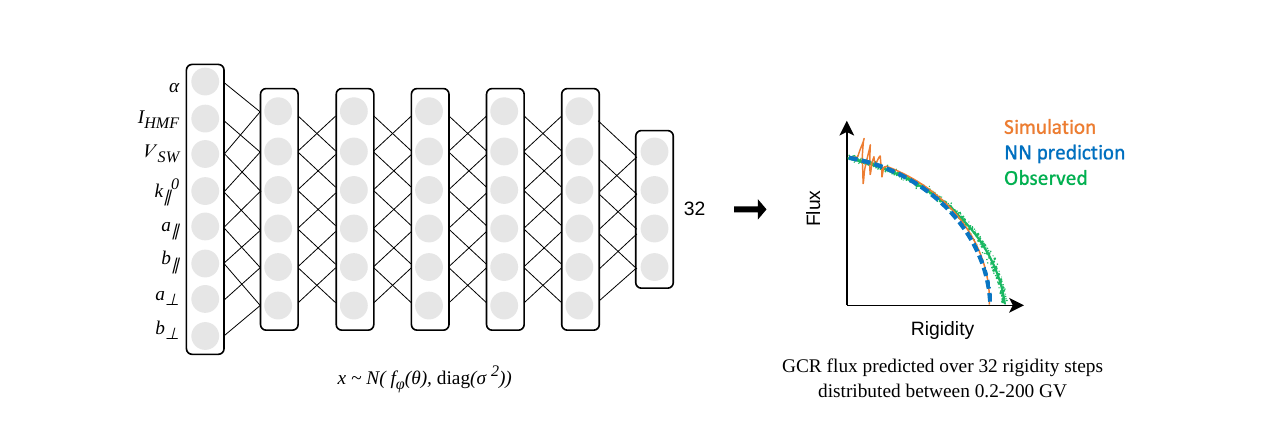}}
   \caption{\change{The NLE (left) takes as input eight latent parameters of the heliosphere and predicts the GCR flux near Earth for 32 rigidity steps. These steps are uniformly distributed in logspace between 0.2 and 200 GV. The NLE is trained with targets from a simulator $S$ that suffers from instabilities (the jagged portion of simulation curve).}{A NN $f_\phi$ (left) is trained to approximate the output of simulator $S$. The NN takes as input eight latent parameters of the heliosphere $\mathbf{\theta} = \{ \alpha, I_{HMF}, V_{SW}, k^{0}{\parallel}, a_{\parallel}, b_{\parallel}, a_{\perp}, b_{\perp} \}$ and deterministically predicts the GCR flux near Earth for 32 rigidity steps (right). These steps are uniformly distributed in logspace between 0.2 and 200 GV. $S$ suffers from instabilities (the jagged portion of simulation curve). The NLE consists of $f_\phi$ combined with a rigidity-dependent Gaussian noise model (Equation~\ref{eq:likelihood}).}}
   \label{fig:nle_diagram}
\end{figure}




The surrogate likelihood function consists of an NN that predicts the observed GCR fluxes over 32 rigidity bins for a given set of eight parameters (Figure~\ref{fig:nle_diagram}) and a rigidity-dependent model of the observation error. This surrogate likelihood can be evaluated quickly at inference time, has an inductive bias towards smoothness, and is differentiable. This conveniently solves all three obstacles to the use of HMC for Bayesian inference.

To train the NN, numerical solutions were computed for almost three million parameter values, similar to what was done in \cite{bib:corti19:numericalmodeling}. This initial computational cost is expensive but embarrassingly parallelizable, so it can be done on a large computing cluster. Solutions that demonstrated numerical instability were removed, resulting in 1,987,658 negative HMF polarity samples, which were further split into training sets (90\%) and test sets (10\%), with the latter used for early stopping (monitored test loss) and hyperparameter optimization via Optuna \cite{optuna}.

The NLE consists of a fully connected network with 8 inputs, five hidden layers of 1024 units each, and 32 outputs. It predicts the modulated flux at 32 rigidity steps, uniformly distributed in logspace between 0.2 and 200 GV, from the 8 input parameters (tilt angle $\alpha$, HMF intensity near Earth $I_{HMF}$, solar wind speed $V_{sw}$, and parameters related to the parallel and perpendicular diffusion coefficients: $k^{0}_{\parallel}$, $a_{\parallel}, b_{\parallel}, a_{\perp}$, and $ b_{\perp}$). Hidden units used the \change{SELU}{scaled exponential linear unit (SELU)} activation function\add{~\cite{klambauer2017self}},

\protect \begin{align}
SELU(x) &= \lambda \begin{cases} 
x & \text{if } x > 0 \\ 
\alpha(e^x - 1) & \text{if } x \leq 0 
\end{cases} \\
\lambda &= 1.0507 \quad \text{(default)}  \nonumber \\
\alpha &= 1.6733 \quad \text{(default).} \nonumber 
\end{align}

\begin{figure}
   \makebox[1\textwidth][c]{\includegraphics[width=0.7\textwidth, trim=0cm 0cm 0cm 1cm, clip]{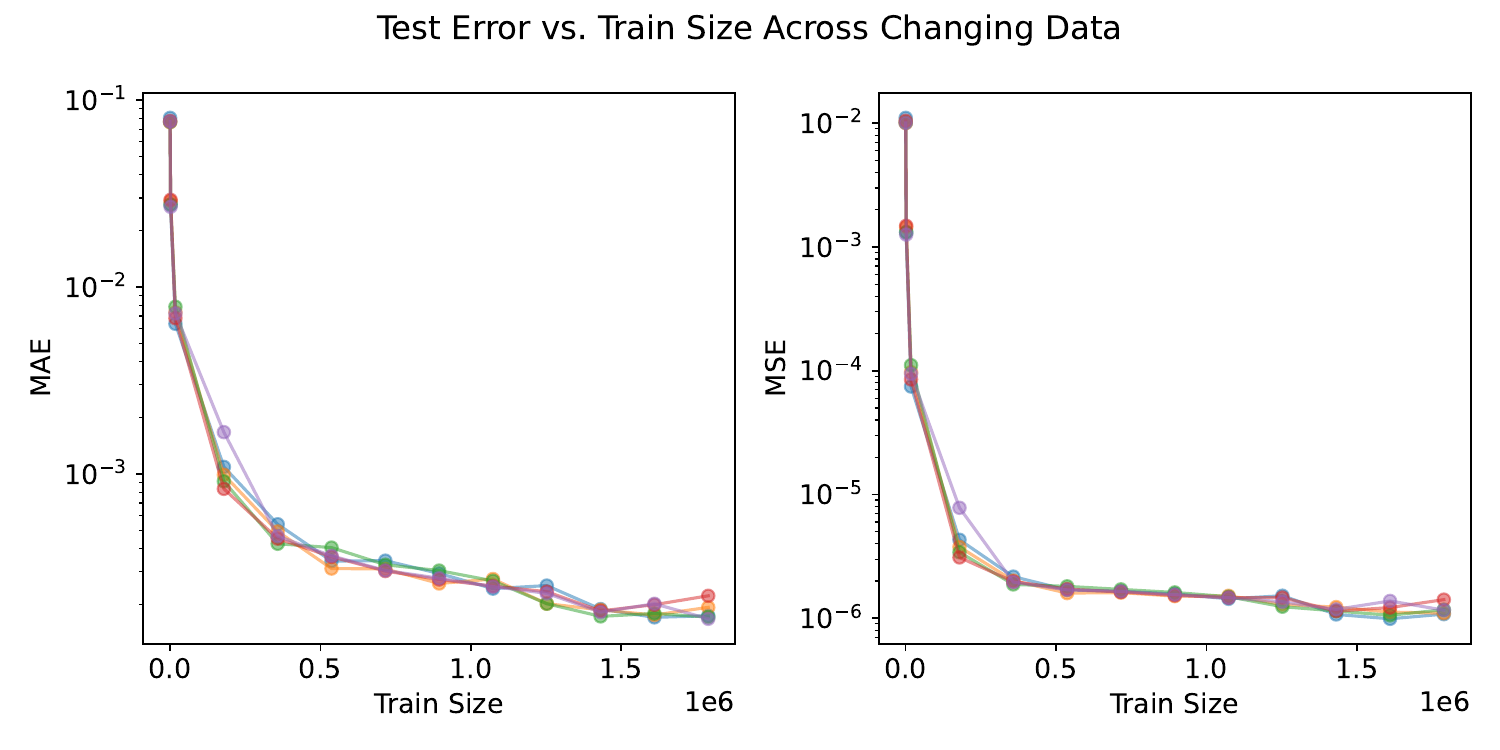}}
   \caption{Mean absolute error (left) and mean squared error (right) with numerical model solutions on the test dataset, shown for five models trained on different bootstrap-sampled datasets.}
   \label{fig:test_error_data}
\end{figure}

\noindent
The mean absolute error was minimized using the Adam optimizer~\cite{kingma2017adam} with the ReduceLROnPlateau learning rate schedule (initial learning rate $2.5 \cdot 10^{-5}$) and a batch size of 128. L2 regularization is used to prevent overfitting (weight $2.6 \cdot 10^{-9}$). 
The relative error of the NLE predictions compared to the held-out numerically stable model solutions was found to be very low with no significant correlations with the input values, as shown in Figure~\ref{fig:test_error_data}. 


\subsection{MCMC Setup}
\label{sec:mcmc}

With our trained NLE, we then perform posterior sampling for $N = 133$ separate observation time intervals: 50 intervals observed by AMS-02 and 83 intervals observed by PAMELA. That is, for each observation $x_n$, $n\in 1 \ldots N$, we perform inference to obtain samples from $p(\theta_n|x_n)$. HMC is performed using the No-U-Turn Sampler~\cite{hoffman2011} and the Dual Averaging Step Size Adaptation kernel of the TensorFlow Probability MCMC package to generate $110,000$ samples from the posterior for each time interval. Hyperparameters are manually adjusted to achieve low autocorrelation between samples, and are set to be 8,000 adaptation steps, maximum tree depth of 10, step size 0.1, maximum energy difference of 1,000, unrolled leapfrog steps of 1, and a target acceptance probability of 75\%. Additionally, we apply thinning by only selecting every other sample and discarding the first 10,000 samples as part of the burn-in process. Thus, the total HMC iterations for each observation is $110,000 \cdot2 + 10,000 = 230,000$, including the burn-in phase and thinning.

Since the diffusion coefficients cannot be directly measured in space, we sample the five parameters related to them: $k^{0}_{\parallel}$, $a_{\parallel}, b_{\parallel}, a_{\perp}$, and $ b_{\perp}$.
On the other hand, the tilt angle, HMF intensity near Earth, and solar wind speed ($\alpha, I_{HMF}$, and $V_{SW}$) are routinely measured by various instruments, and so they are kept constant during the inference process, with values fixed to their 1-year backward average for each interval, using data from NASA/GSFC's OMNI daily dataset \cite{bib:omni} via OMNIWeb (\url{https://omniweb.gsfc.nasa.gov/}) and from the Wilcox Solar Observatory (\url{http://wso.stanford.edu/Tilts.html}). 




The NLE is trained on data from a limited domain of parameter space and so should not be expected to generalize well outside this domain. We prevent the HMC from sampling outside the ``trusted'' domain with the choice of prior distribution, $p(\theta)$. This prior is uniform in the domain of the training data \change{and decays exponentially (rate $\lambda = 10^6$) in every direction outside the domain}{and decays exponentially in every direction outside the domain (with a rate of $\lambda = 10^6$ in the standardized parameter space)}.

\subsection{Assessing Convergence}
\label{sec:assess_convergence}

As described in Section~\ref{sec:evaluating-the-results}, when using an NLE to approximate the likelihood, one should empirically validate the results. Each of the three steps described in Figure~\ref{fig:diagram} can lead to errors in the predicted posterior: sampling a finite training dataset, initializing and training the NLE, and sampling from the posterior with HMC. To ensure that our posterior inference is accurate, we use two types of empirical evaluations: a posterior predictive test on held-out simulation data, and the Gelman-Rubin statistic on bootstrapped experiment replications.

The Gelman-Rubin $\hat{R}$ statistic~\cite{Gelman2013Bayesian} is widely used to assess MCMC convergence. MCMC is run with different initializations, and $\hat{R}$ indicates whether the resulting empirical distributions are different. For our SBI use case, we propose to instead compare replications of the complete inference procedure to account for the additional opportunities for modeling errors:

\begin{enumerate}
    \item State initializations for the MCMC (HMC).
    \item Initializations and training of the NLE (model).
    \item Sampling a finite training dataset used for NLE training (data).
\end{enumerate}

In our experiments, we accounted for all three of these sources of variability, using bootstrap sampling to approximate the last. Five experimental replications were performed and used to compute a Gelman-Rubin $\hat{R}$ statistic.  If the posterior samples are similar between all three of these chains ($\hat{R} < 1.1$ in our application), then we consider the posterior samples obtained with the surrogate NLE to have converged. 

\section{Results}
\label{sec:results}

\subsection{Single Interval Convergence}

In this section, we focus on a single time interval (one of $N$ inference tasks) to demonstrate our analysis. We also compare the Gelman-Rubin $\hat{R}$ statistic obtained when using the three different types of experiment replication. From the weakest to strongest, we refer to these as HMC replicates, model replicates, and data replicates. Each type of replicate subsumes the previous types, so data replicates also use different neural network and MCMC state initializations.

\begin{figure}[!h]
   \makebox[1\textwidth][c]{\includegraphics[width=1.3\textwidth, trim=0cm 0cm 0cm 1.5cm, clip]{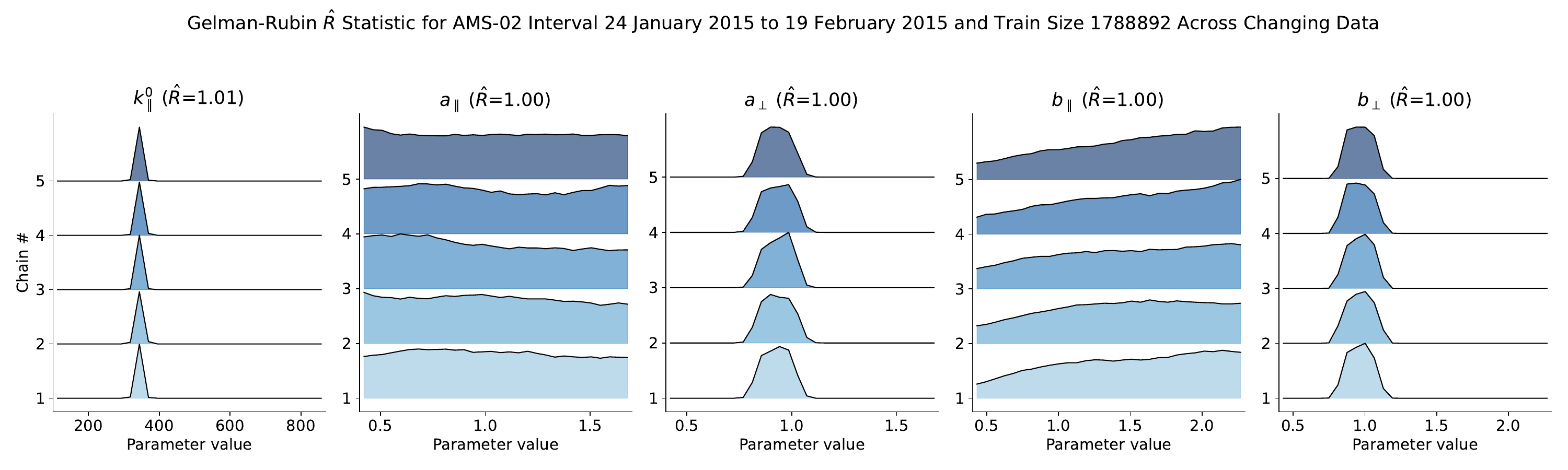}}
   \caption{Posterior samples for AMS-02 interval 24 January to 19 February 2015 across the five sampled parameters. Plotted histograms show the sample distribution from five data replicates of full train size (NLE trained on different bootstrap-sampled training datasets). The Gelman-Rubin $\hat{R}$ statistic for each parameter is shown. Histograms are scaled to the same height across parameters.}
   \label{fig:single_interval_hist_ams}
\end{figure}

Figure~\ref{fig:single_interval_hist_ams} shows the empirical marginal distributions across five data replicates. Each replicate uses a different bootstrap-sampled dataset (1,788,892 examples) for training the NLE (same architecture but different initialization). The replicates exhibit similar distributions and the $\hat{R}$ values are below the threshold of 1.1 for each of the five parameters. The largest differences between replicates are seen in the parallel diffusion coefficient parameters, which are less constrained by the data; this is expected, since perpendicular diffusion dominates the transport processes in the majority of the heliosphere. 

\begin{figure}[!h]
   \makebox[1\textwidth][c]{\includegraphics[width=1.3\textwidth, trim=0cm 0cm 0cm 1.5cm, clip]{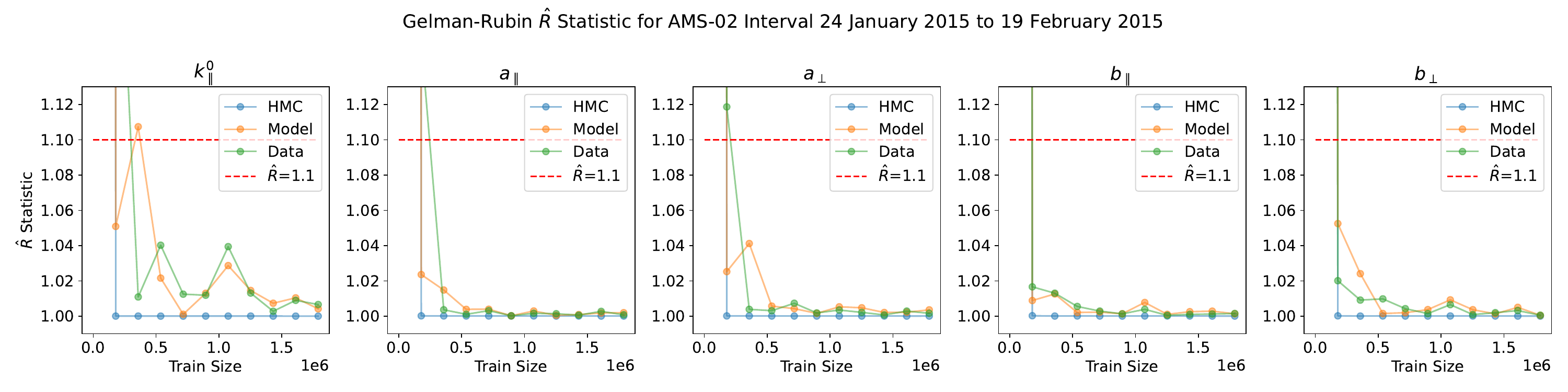}}
   \caption{Gelman-Rubin $\hat{R}$ statistic for AMS-02 interval 24 January to 19 February 2015 plotted for the five sampled parameters against training set size. Lines show the statistic across five HMC replicates (blue), model replicates (orange), and data replicates (green). The dashed red line marks the convergence threshold of $\hat{R} = 1.1$.}
   \label{fig:single_interval_rhat_ams}
\end{figure}

We next evaluated the choice of experiment replication method by running five replicates from each method and comparing the Gelman-Rubin $\hat{R}$s. We expect any difference in these methods to disappear as the size of the training dataset increases, so we repeated the experiment for different-sized subsets of the training data (i.e. simulating a smaller training set). The results in Figure~\ref{fig:single_interval_rhat_ams} show that the data and model replicates are significantly more variable than the HMC replicates, having a higher value of $\hat{R}$ for most training set sizes, especially for smaller training sets. Even in this example, there are scenarios where the model and/or data replicate is above the $\hat{R}=1.1$ threshold while the HMC and model replicates are below it. This strongly suggests that running data replicates should be a standard evaluation protocol to ensure that the posterior obtained from the neural surrogate HMC is repeatable. 

\subsection{Convergence Across All Intervals}

\begin{figure}[!h]
   \makebox[1\textwidth][c]{\includegraphics[width=1.3\textwidth, trim=0cm 0cm 0cm 1.5cm, clip]{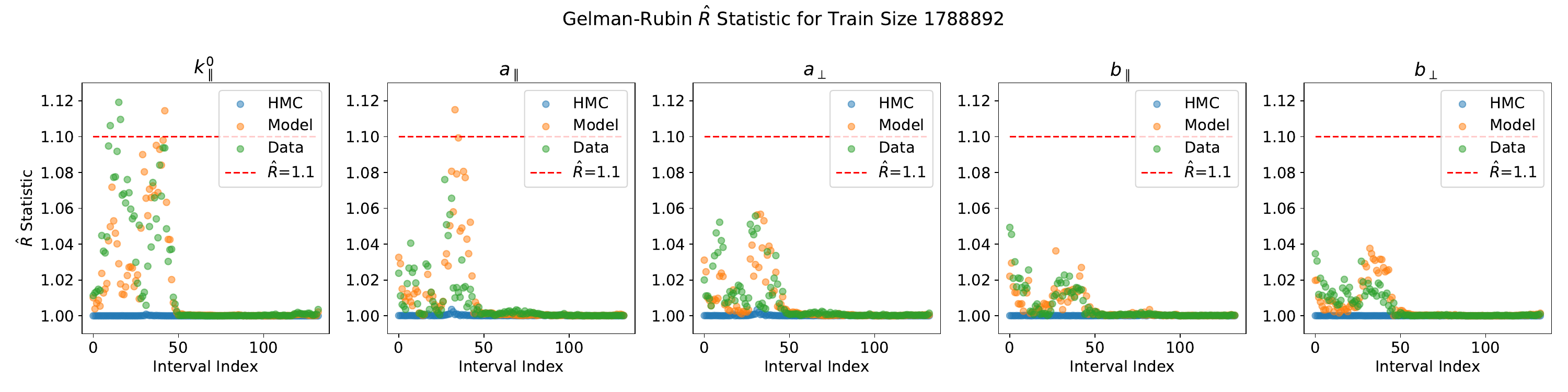}}
   \caption{Gelman-Rubin $\hat{R}$ statistic for the full training dataset size plotted for the five sampled parameters against time interval index. Dots show the statistic across five HMC replicates (blue), model replicates (orange), and data replicates (green). The dashed red line marks the convergence threshold of $\hat{R} = 1.1$. Intervals 0-49 are observations taken by AMS-02, and intervals 50-132 are observations taken by PAMELA.}
   \label{fig:all_intervals_single_train_rhat}
\end{figure}

The experiments above were repeated for each of the $N=133$ negative polarity intervals. The $\hat{R}$ statistics for the full training dataset are plotted in Figure~\ref{fig:all_intervals_single_train_rhat}. Intervals 0-49 are posterior samples for observations from AMS-02, and intervals 50-132 are observed by PAMELA.

It can be seen that for chains with different model initializations and training data, there is a large difference between posterior samples for AMS-02 and PAMELA observations. We suspect this difference is primarily due to the difference in experimental uncertainty values between AMS-02 and PAMELA. AMS-02 has much smaller uncertainties (roughly an order of magnitude) than PAMELA, leading to a higher sampling sensitivity to changes in model initialization and training data. This higher sensitivity is also reflected in the width of the PDFs, which are two to three times narrower for $k^{0}_{\parallel}$ and $b_{\perp}$ when using AMS-02 instead of PAMELA data (compare Figures~\ref{fig:single_interval_hist_ams} and~\ref{fig:single_interval_hist_pamela}). In general, PAMELA posterior samples exhibit a stronger posterior convergence for all train sizes when compared to AMS-02 intervals, as seen in Figure~\ref{fig:all_intervals_single_train_rhat} and the differences between Figures~\ref{fig:single_interval_hist_ams} and~\ref{fig:single_interval_hist_pamela}.

\begin{figure}[!h]
   \makebox[1\textwidth][c]{\includegraphics[width=1.3\textwidth, trim=0cm 0cm 0cm 1.5cm, clip]{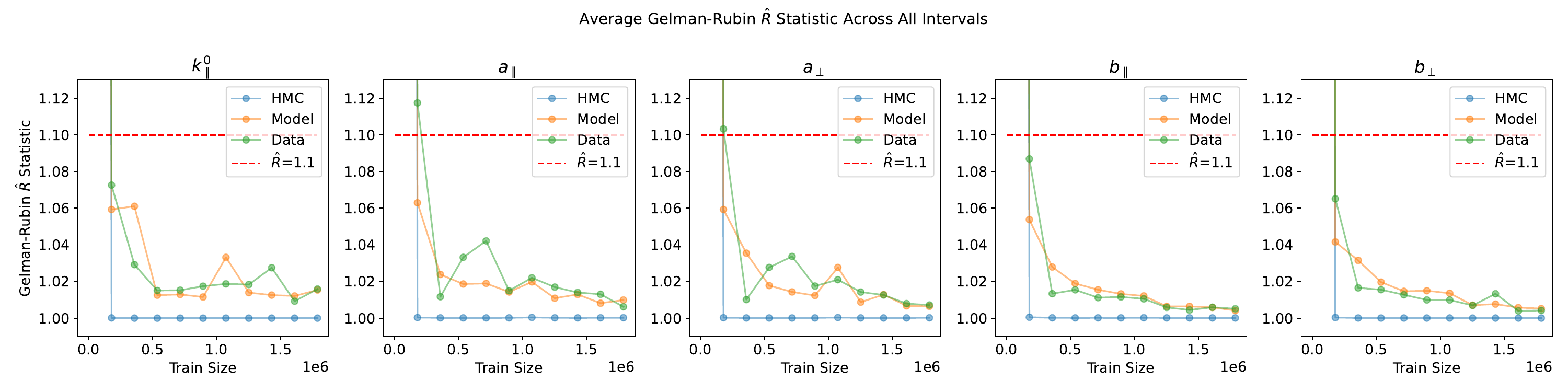}}
   \caption{Gelman-Rubin $\hat{R}$ statistic averaged across all time intervals plotted for the five sampled parameters against training set size. Lines show the statistic across five HMC replicates (blue), model replicates (orange), and data replicates (green). The dashed red line marks the convergence threshold of $\hat{R} = 1.1$.}
   \label{fig:all_intervals_all_train_rhat}
\end{figure}

To examine the effect of training set size on posterior convergence, the $\hat{R}$ statistics were averaged across all 133 time intervals (Figure~\ref{fig:all_intervals_all_train_rhat}). This plot confirms that the model and data replicates show significantly more variability, and that the data replicate should be preferred in practice.

\subsection{Posterior Predictive Test}

\begin{figure}[!h]
   \centering
   \makebox[1\textwidth][c]{\includegraphics[width=1.3\textwidth, trim=0cm 0cm 0cm 1.5cm, clip]{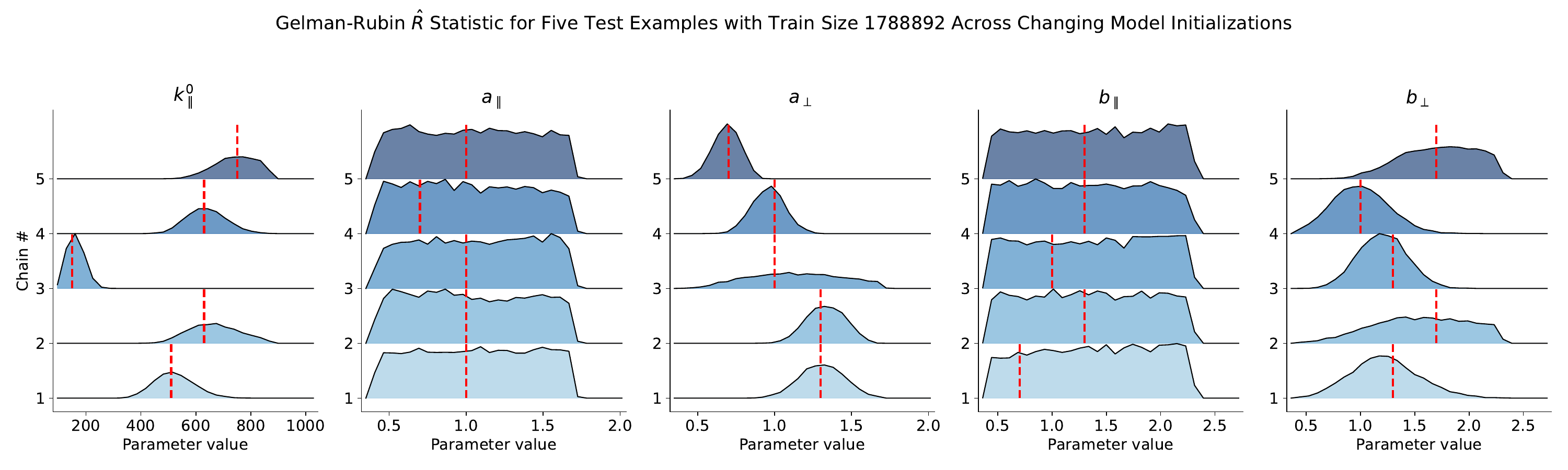}}
   \caption{Marginal posterior distributions for five typical simulations, with the true parameters shown as red dashed lines. The NLE was trained on the full training set.}
   \label{fig:ppt_one_example}
\end{figure}

Consistent posteriors between replications do not imply that the procedure has converged to the true posterior --- it is possible that the inductive bias of the NLE model gives a consistently biased posterior. Thus, we also recommend an empirical evaluation that the posterior is correct. This can be done by testing that the predicted posteriors are well-calibrated on simulated observations (for which we have the true latent parameter value). Because inference is computationally expensive, here we perform posterior predictions on a held-out set of 100 simulations and verify that the true parameter is included in the marginal posteriors. Figure~\ref{fig:ppt_one_example} shows five typical examples; the true parameter values are all in high-likelihood regions of the marginal distributions. 

\begin{figure}[!h]
   \makebox[1\textwidth][c]{\includegraphics[width=1.3\textwidth, trim=0cm 0cm 0cm 1.5cm, clip]{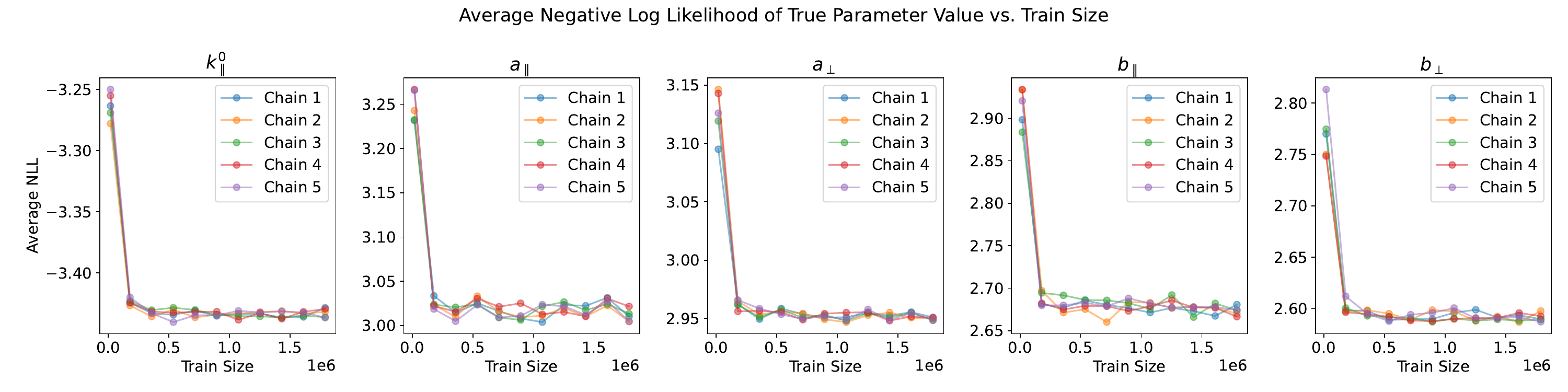}}
   \caption{Average NLL of the true parameters over 100 held-out simulations, using the NLL calculation described in the text. The five data replicates all quickly converge to the same NLL.}
   \label{fig:ppt_all_test_examples}
\end{figure}

This is quantified in Figure~\ref{fig:ppt_all_test_examples}, where the empirical histograms of the marginals with a fixed set of 30 bins are used to define categorical distributions and a log likelihood. The negative log likelihood (NLL) of the true parameter is computed for each prediction and averaged across the 100 held-out simulations. The NLL is plotted for five data replicates with different training dataset sizes, which all quickly converge to the same NLL value. The small variation in the NLL across train set sizes of Figure~\ref{fig:ppt_all_test_examples} and low $\hat{R}$ statistic of Figure~\ref{fig:all_intervals_all_train_rhat} indicates that even for a small number of training examples ($\approx10\%$ of our entire training set), the posterior samples converge to the correct parameter and are consistent between HMC, model, and data replicates.

\subsection{Choice of Prior and Observational Noise Model}

\begin{figure}
   \centering
   \makebox[1\textwidth][c]{\includegraphics[width=1.3\textwidth, trim=0cm 0cm 0cm 2cm, clip]{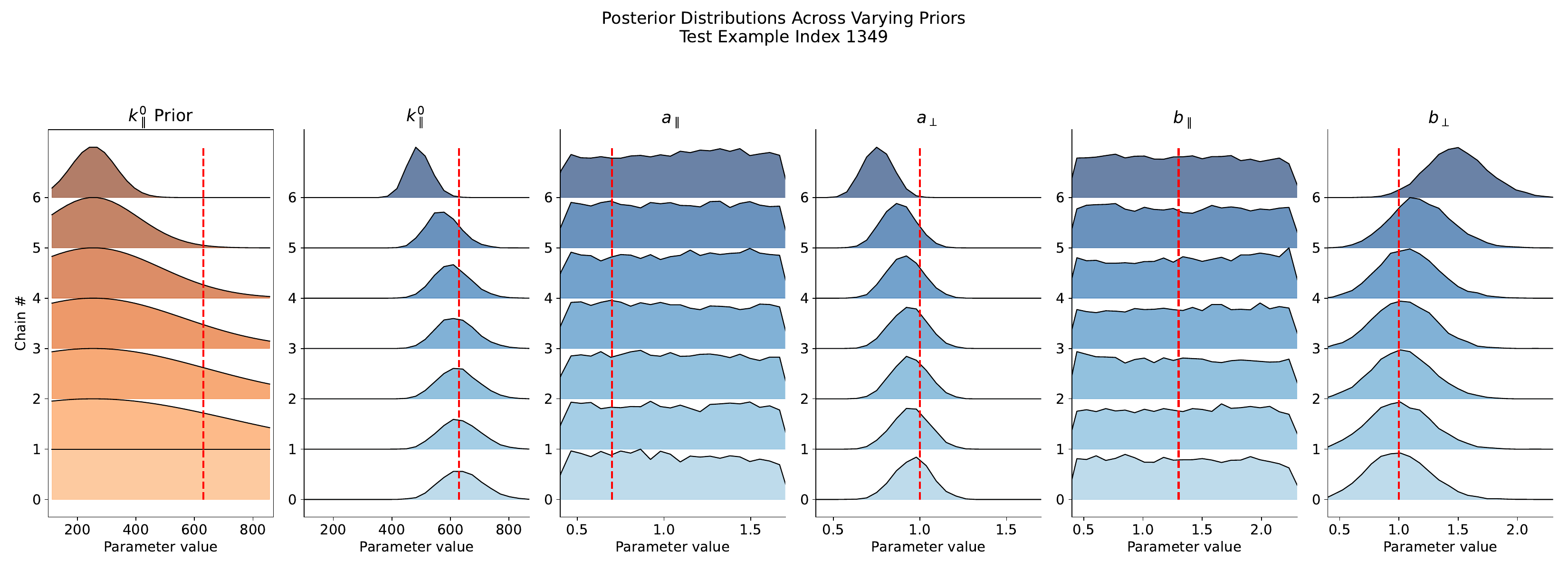}}
   \caption{Marginal posterior distributions for a typical simulation, with the true parameter value shown as a red dashed line. The leftmost column shows the choice of prior over parameter $k^{0}_{\parallel}$ (all other parameters have a uniform prior). \add{The worst prior is shown in the top row, with subsequent rows showing better priors. The bottom row corresponds to the reference (uniform) prior. }The NLE was trained on the full training set.}
   \label{fig:prior_ppt}
\end{figure}

Further experiments were performed to demonstrate the robustness of our approach to different priors and noise distributions. Figure~\ref{fig:prior_ppt} demonstrates how the inference procedure performs well even with a bad prior (top row) --- as the prior distribution improves and the true parameter value has a higher likelihood under the prior distribution, the marginal posterior is better able to capture the true value.

Our approach also generalizes to different noise models. We demonstrated this by replacing the Gaussian observational noise model with a heavy-tailed distribution\add{ for an example interval}. The result is a relaxation of the constraints on the model parameters and more disperse marginal posteriors (see \change{Figure~\ref{fig:noise_model_ams02}}{\ref{app:noise}}).\add{ More complex noise models could include correlated noise across rigidity bins (see, e.g., \citeA{bib:derome19:fitting-cosmicray-data} and \citeA{bib:cuoco19:scrutinizing-evidence-dark}), which we would expect to further loosen the constraints on the parameters.}



\subsection{Computational Speedup of the NLE}

To provide a concrete demonstration of the computational advantage of our method, we also experimented with using a traditional Bayesian inference method. Using the simulator to compute the likelihoods, the no-gradient Random Walk Metropolis-Hastings algorithm (RWMH)~\cite{metropolis1953} was used to generate $400,000$ samples over each time interval. The scale of the normally distributed perturbations during RWMH was manually set to 0.01 to achieve an acceptance ratio close to 25\%. We use the same thinning and number of burn-in steps as HMC, bringing the total number of RWMH iterations for each observation to $400,000 \cdot 2 + 10,000 = 810,000$. 

\begin{figure}
   \centering
   \makebox[1\textwidth][c]{\includegraphics[width=1.3\textwidth]{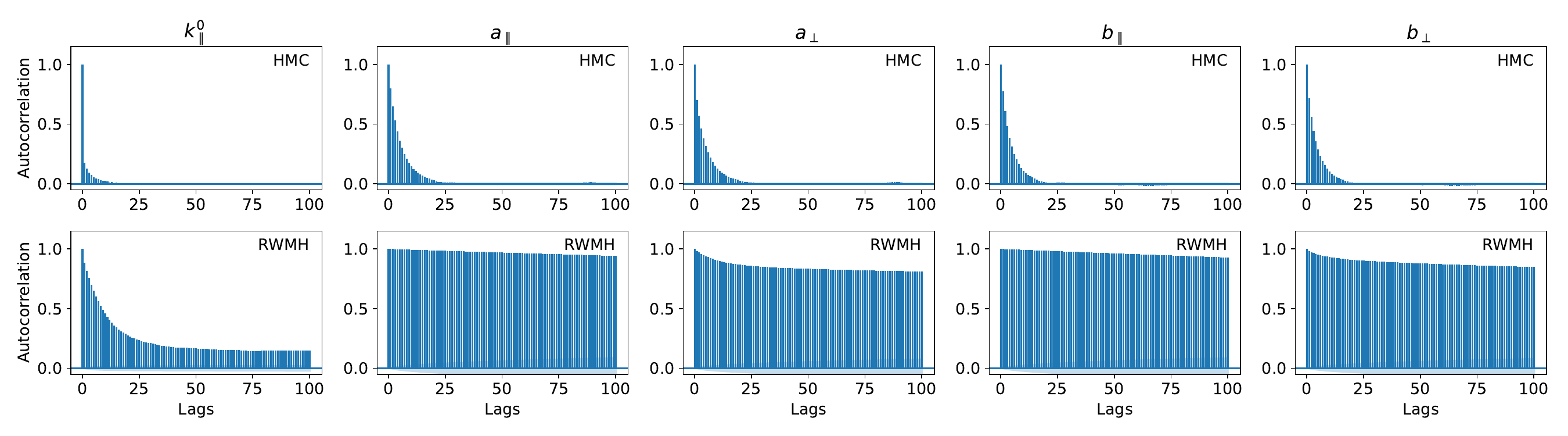}}
   \caption{
      Comparison of autocorrelation in HMC (top) and RWMH (bottom) samples for all sampled parameters. Both HMC and RWMH samples were generated with the hyperparameters described in Section~\ref{sec:mcmc} and a thinning value of 1. The data used to compute the likelihood of samples were measured by AMS-02 in the negative polarity time interval 24 January to 19 February 2015, and the NLE used to compute likelihood evaluations was trained on the full training set.
    }
   \label{fig:acf_comparison}
\end{figure}

Figure~\ref{fig:acf_comparison} shows that the autocorrelation of these samples is much higher than that of HMC. The RWMH samples would need to be thinned by at least a factor of $10^2$ to achieve the same autocorrelation as HMC, so obtaining $10^5$ uncorrelated samples with RWMH for a single observation would require at least $10^7$ likelihood evaluations. This is already a 10-fold increase over our NLE approach (which required $10^6$ likelihood evaluations for training). Since the NLE can be reused for all 133 observations, the savings become three orders of magnitude. Furthermore, the speedup at inference time is immense: on a 1-core CPU the simulator takes $\approx$ 5m50s~\cite{bib:corti19:numericalmodeling} and often fails due to numerical instabilities, whereas the NLE is stable and has an inference time of $\approx21$ms. This means that, even assuming the best case of no numerical instabilities, the cost of generating the training data set for the NLE ($2 \cdot 10^6$ numerical model evaluations, for a total of $\approx1.9 \cdot 10^5$ CPU hours) is amortized after \change{generating $10^5$ uncorrelated HMC samples with simulator $\mathcal{S}$ for 9 observations ($2 \cdot 10^6 < 2.4 \cdot 10^5 \cdot 9$).}{performing inference (generating $10^5$ samples) for 9 observations ($2 \cdot 10^6 < 2.3 \cdot 10^5 \cdot 9$).}

\subsection{GCR Transport Parameters}

\begin{figure*}
   \centering
   \includegraphics[width=\textwidth]{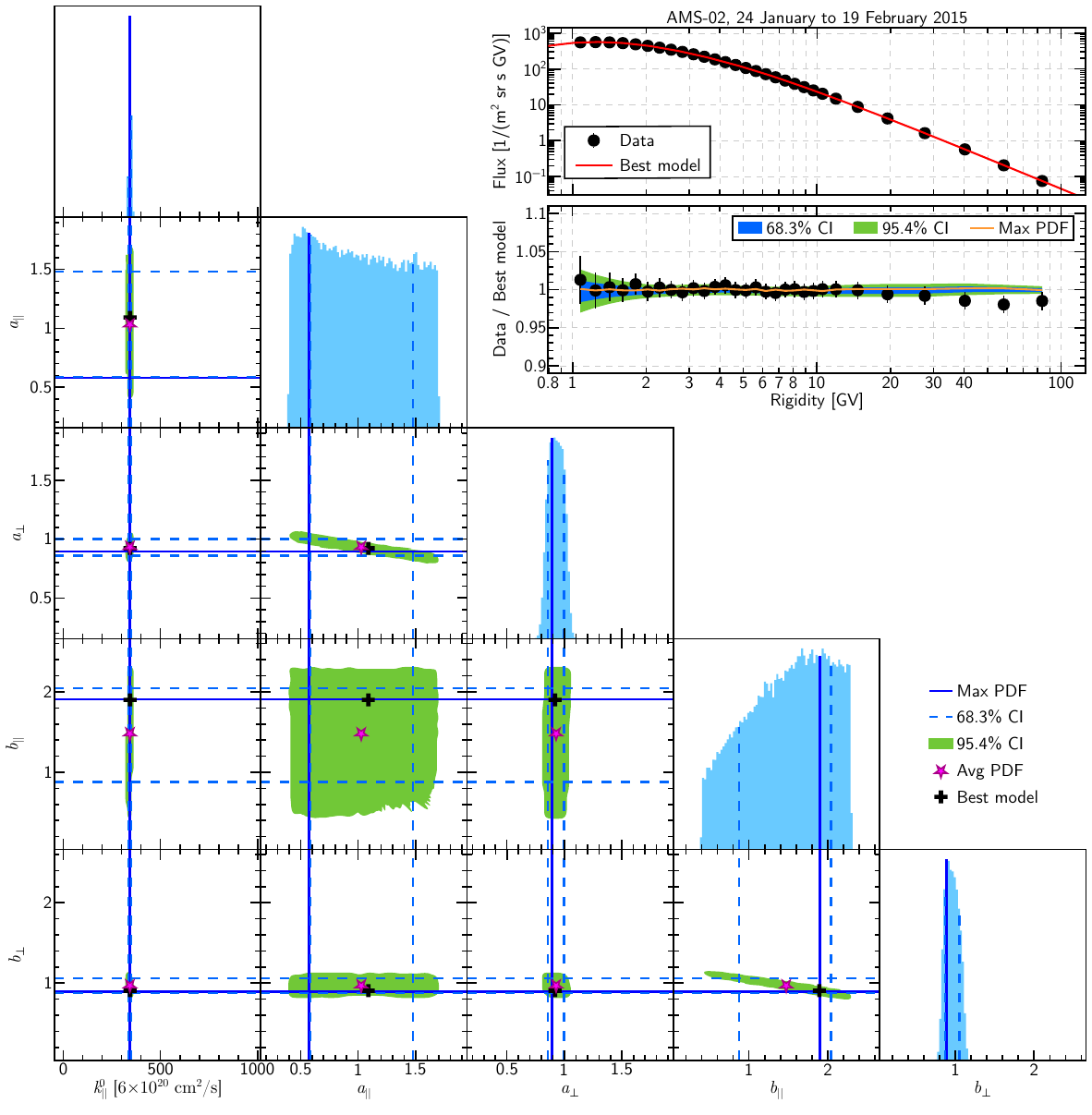}
   \caption{
      2D and 1D PDFs of the diffusion coefficient parameters obtained from the HMC for GCR flux measured by AMS-02 in the negative polarity time interval 24 January to 19 February 2015. The azure area is the 1D marginalized PDF over each parameter, and the magenta star and dark blue line show the average and maximum of that 1D PDF, respectively. The blue dashed lines \change{shows}{show} the 1D 68.3\% credible intervals, and the green area shows the 2D 95.4\% credible intervals. The black cross shows the parameter values corresponding to the maximum likelihood in the 5D PDF. The top right panel shows a comparison of the maximum likelihood NLE (red line) with observations, together with the 68\% and 95\% credible intervals.
   }
   \label{fig:coefficient_pdfs_ams}
\end{figure*}

The converged posterior samples provide state-of-the-art constraints on parameter PDFs for AMS-02 and PAMELA observations. An example of the PDFs of the free parameters obtained with HMC for a single experiment using AMS-02 data is shown in Figure~\ref{fig:coefficient_pdfs_ams} (see Figure~\ref{fig:coefficient_pdfs_pamela} in the Appendix for an example using PAMELA data). The NLE used to generate these results was trained on the full dataset as described in Section~\ref{sec:nle_setup} without bootstrap sampling. The normalization of the diffusion coefficient ($k^{0}_{\parallel}$) and the slopes of the perpendicular diffusion coefficient ($a_{\perp}$ and $b_{\perp}$) are well constrained by the data, as shown by their narrow PDFs. The slopes of the parallel diffusion coefficient ($a_{\parallel}$ and $b_{\parallel}$) are less constrained, as expected, since perpendicular diffusion dominates the transport processes in the majority of the heliosphere. \remove{These results are in agreement with what was found in \cite{bib:corti19:numericalmodeling} using a grid-based ordinary least-squares minimization procedure on the same AMS-02 data and numerical model. We improve on \cite{bib:corti19:numericalmodeling} by smoothing the numerical instabilities of the model and by providing posterior PDFs for all free parameters of the numerical model instead of point estimates. In addition, we reduce the month-to-month scattering and uncertainty on the well-constrained parameters ($k^{0}_{\parallel}$, $a_{\perp}$, and $b_{\perp}$), produce more robust uncertainties for the least-constrained parameters ($a_{\parallel}$ and $b_{\parallel}$), and find closer model solutions to the data.}

\begin{figure*}
   \centering
   \includegraphics[width=\textwidth]{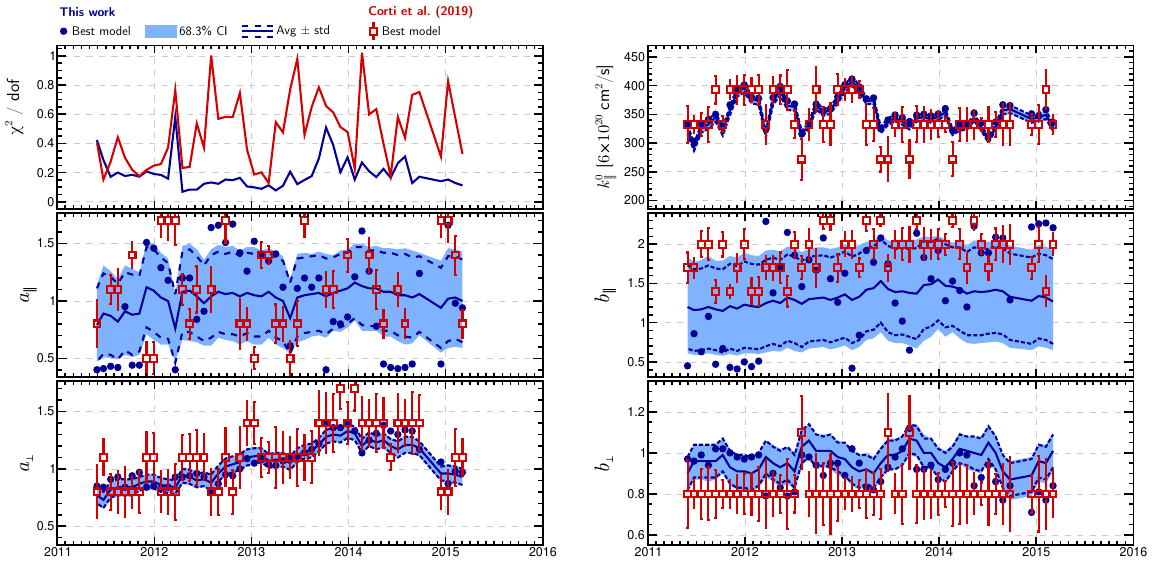}
   \caption{\add{Comparison of the parameters inferred by our method (blue) versus those inferred in \citeA{bib:corti19:numericalmodeling} (red) for each time interval. Solid blue lines plot the mean posterior, with dashed lines at $\pm$ one standard deviation of the 1D marginalized PDF for each parameter. The blue shaded region indicates the 68.3\% credible interval (the region between the 0.159 and 0.841 quantiles). The blue circles indicate the \textit{maximum a posteriori} (MAP) estimates of the joint PDFs (in 5D). The red squares are the maximum likelihood point estimates from \citeA{bib:corti19:numericalmodeling}, with the error bars representing frequentist 68\% confidence intervals.}
   }
   \label{fig:comparison}
\end{figure*}

\add{These results improve upon those obtained in \citeA{bib:corti19:numericalmodeling}, which used a grid-based ordinary least-squares minimization procedure on the same AMS-02 data and numerical model. Figure~\ref{fig:comparison} shows the differences in the inference results. The improvements come from smoothing the numerical instabilities of the model, exploring the continuous parameter space rather than relying on a coarse grid, and providing posterior PDFs for all free parameters of the numerical model in a Bayesian manner instead of computing maximum likelihood point estimates and using frequentist confidence intervals to estimate uncertainty. Specifically, we observe that:}

\begin{enumerate}
    \item \add{The uncertainties on the well-constrained parameters were overestimated in \citeA{bib:corti19:numericalmodeling} by a factor of 1.5 to 4.}
    \item \add{The uncertainties on the least-constrained parameters ( $a_{\parallel}$, and $b_{\parallel}$) were underestimated in \citeA{bib:corti19:numericalmodeling} by a factor of 2 to 5.}
    \item \add{The $\chi^2$ values divided by the degrees of freedom are smaller for our method compared to the grid-based method, corresponding to higher likelihoods.}
\end{enumerate}

The posterior samples over the free parameters obtained with HMC for all 133 time intervals analyzed in this paper are publicly available, and instructions for their use are detailed in the Open Research Section. The physical interpretation of the results obtained with the HMC analysis will be the subject of a future publication.

\section{Conclusions}

We have demonstrated that neural surrogate HMC is a practical method for solving certain Bayesian inference problems, and we provide specific recommendations for its application. Our full contributions are:

\begin{enumerate}
    \item We characterized the three key advantages of neural surrogate HMC for SBI --- namely, that a surrogate NLE yields fast, differentiable likelihood evaluations that smooth out numerical instabilities in the simulator~$\mathcal{S}$.
    
    \item We delineated the regimes in which neural surrogate HMC is preferable to other SBI approaches. These include scenarios that require repeated inference, where training a single reusable NLE amortizes dataset-construction costs and enables rapid posterior inference.
    
    \item We outlined practical guidance for specifying a prior and constructing a training dataset for a surrogate NLE, recommending a prior that penalizes beyond the training domain, and a training set that covers the full space of parameter values considered during inference.
    
    \item We established new guidelines for assessing the convergence of a surrogate likelihood. We recommend evaluating convergence with respect to the complete training dataset, NLE initialization, and HMC initialization; we proposed using the Gelman–Rubin $\hat{R}$ statistic on chains that vary systematically across these three axes.
    
    \item We demonstrated state-of-the-art constraints on heliospheric transport of GCRs. The resulting posterior PDFs quantify the uncertainty in each transport parameter and illustrate both the computational speedup and the convergence of our method.
\end{enumerate}


The computational speedup provided by this method enables the analysis of larger datasets. In future work, we intend to infer parameters for GCR observations taken over the course of more than 50 years at various locations across the heliosphere. 





\newpage
\appendix

\section{Example of HMC inference on a PAMELA time interval}\label{app:pamela}

\begin{figure}[!h]
   \makebox[1\textwidth][c]{\includegraphics[width=1.3\textwidth, trim=0cm 0cm 0cm 1.5cm, clip]{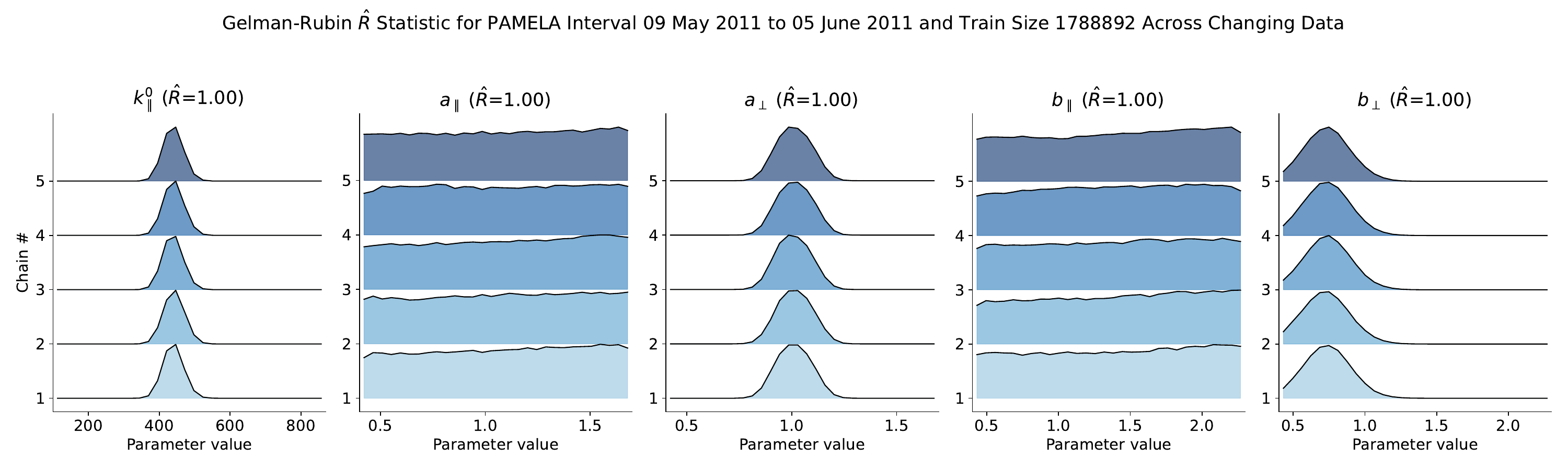}}
   \caption{Posterior samples for PAMELA interval 9 May to 5 June 2011 across the five sampled parameters. Plotted histograms show the sample distribution from five data replicates of full train size (NLE trained on different bootstrap-sampled training datasets). The Gelman-Rubin $\hat{R}$ statistic for each parameter is shown. Histograms are scaled to the same height across parameters.}
   \label{fig:single_interval_hist_pamela}
\end{figure}

\begin{figure}[!h]
   \makebox[1\textwidth][c]{\includegraphics[width=1.3\textwidth, trim=0cm 0cm 0cm 1.5cm, clip]{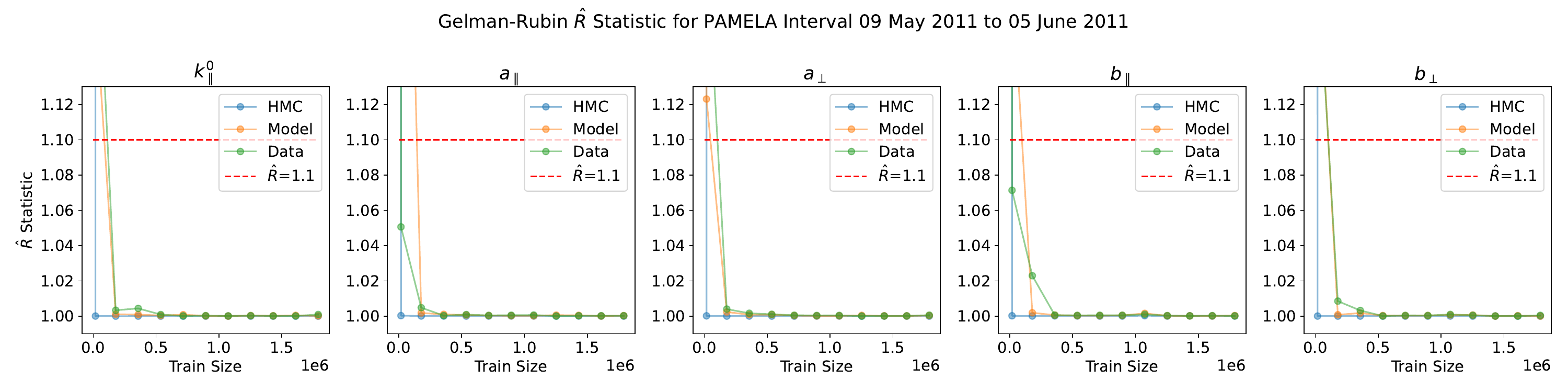}}
   \caption{Gelman-Rubin $\hat{R}$ statistic for PAMELA interval 9 May to 5 June 2011 plotted for the five sampled parameters against training set size. Lines show the statistic across five HMC replicates (blue), model replicates (orange), and data replicates (green). The dashed red line marks the convergence threshold of $\hat{R} = 1.1$.}
   \label{fig:single_interval_rhat_pamela}
\end{figure}

In this section, we present the posterior sampling results of Neural Surrogate HMC for a PAMELA time interval (9 May to 5 June 2011). Due to the comparatively large uncertainties in PAMELA relative to AMS-02, the sampler exhibits low sensitivity to variations in model initialization, training data, and HMC initialization across nearly all training set sizes, as illustrated in Figures \ref{fig:single_interval_hist_pamela} and \ref{fig:single_interval_rhat_pamela}.

\begin{figure*}
   \centering
   \includegraphics[width=\textwidth]{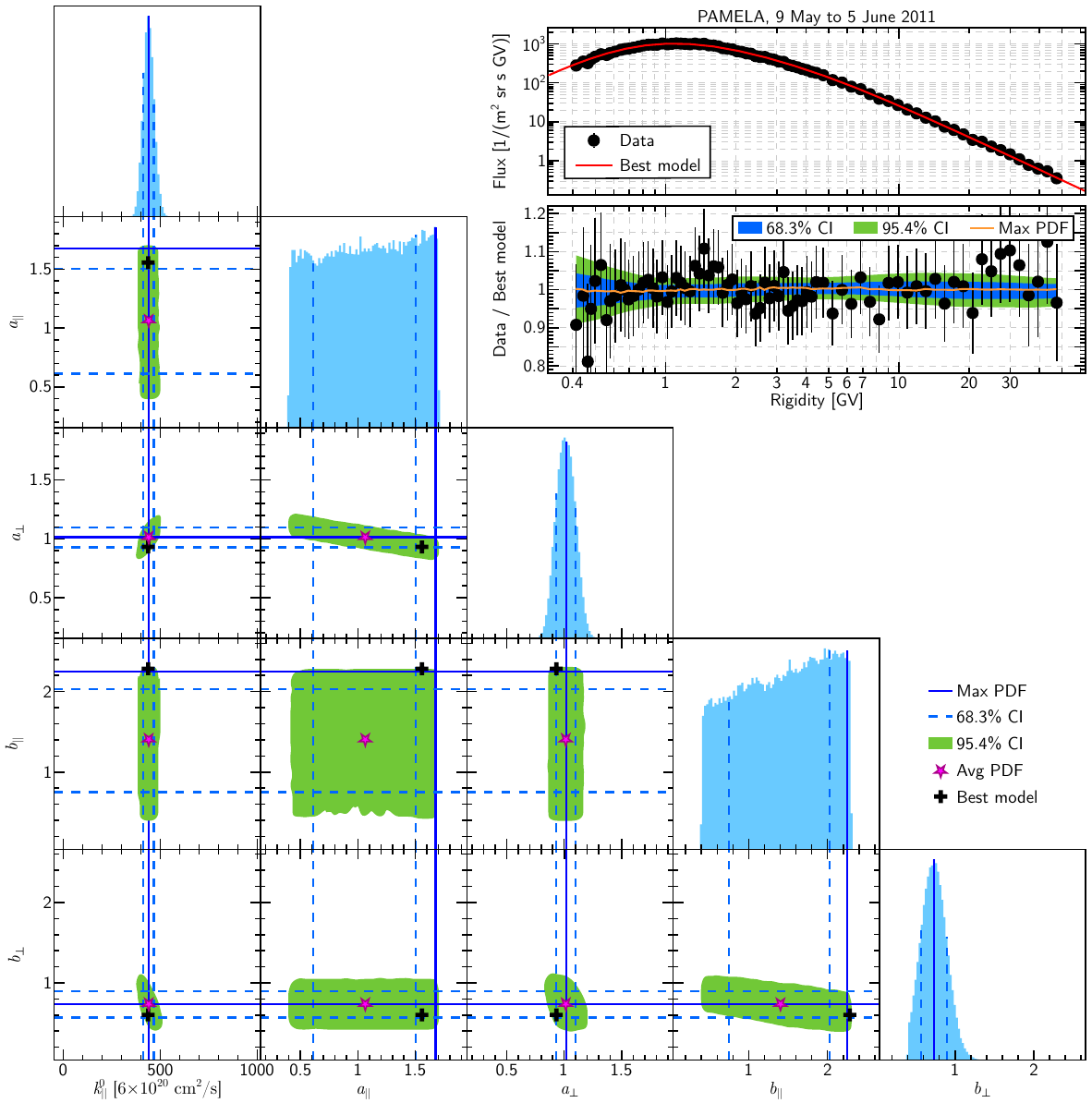}
   \caption{
      2D and 1D PDFs of the diffusion coefficient parameters obtained from the HMC for GCR flux measured by PAMELA in the negative polarity time interval 9 May to 5 June 2011. The azure area is the 1D marginalized PDF over each parameter, and the magenta star and dark blue line show the average and maximum of that 1D PDF, respectively. The blue dashed lines \change{shows}{show} the 1D 68.3\% credible intervals, and the green area shows the 2D 95.4\% credible intervals. The black cross shows the parameter values corresponding to the maximum likelihood in the 5D PDF. The top right panel shows a comparison of the maximum likelihood NLE (red line) with observations, together with the 68\% and 95\% credible intervals.
   }
   \label{fig:coefficient_pdfs_pamela}
\end{figure*}

\section{Heavy-tailed observational noise distribution}\label{app:noise}

\begin{figure}[!h]
   \centering
   \makebox[1\textwidth][c]{\includegraphics[width=1.3\textwidth, trim=0cm 0cm 0cm 2cm, clip]{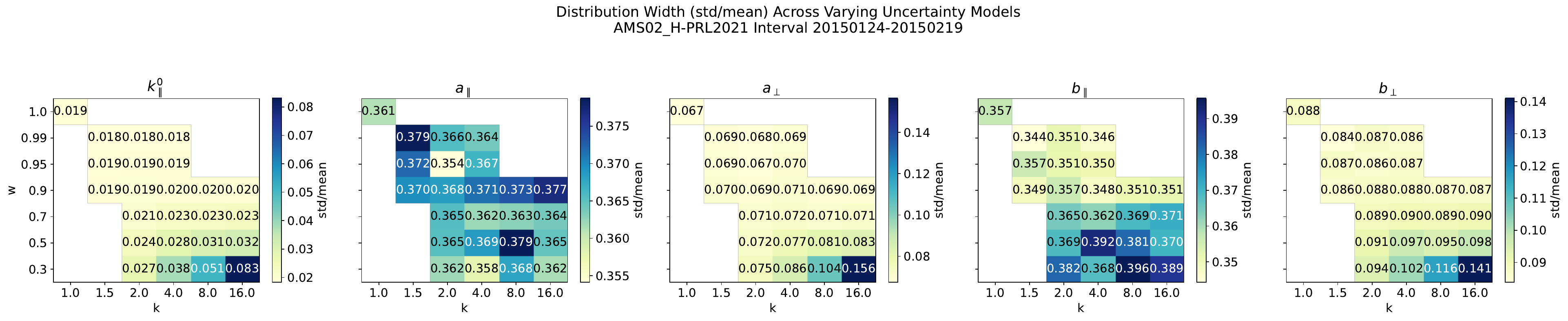}}
   \caption{Marginal posterior distribution width (standard deviation / mean) for shrinking $w$ and growing $k$ values for \change{a single}{an example} AMS-02 interval (AMS-02 interval 24 January to 19 February 2015) against the baseline Gaussian noise model ($w=1$, $k=1$). The NLE was trained on the full training set.\change{ In our original formulation, we assumed that both the statistical uncertainties ($dF_{\text{stat}}$) and the systematic uncertainties ($dF_{\text{syst}}$) were normally distributed and uncorrelated, and thus standard deviation for a given bin $i$ was defined as $\sigma_i = \sqrt{dF_{\text{stat}, i}^2 + dF_{\text{syst}, i}^2}$, leading to a standard $\chi^2$ log-likelihood (Section \ref{sec:description_of_problem}). We now treat the measured flux $F_i$ as a random variable distributed as the convolution of a Normal distribution (representing statistical noise) and a fat-tailed distribution $S$ (representing systematic noise). $S$ is modeled as a Gaussian Mixture Model (GMM) composed of a weighted sum of two Normal distributions with the same mean but different standard deviations: $S(\delta) = w \cdot \mathcal{N}(\delta; 0, dF_{\text{syst}}) + (1-w) \cdot \mathcal{N}(\delta; 0, k \cdot dF_{\text{syst}})$ where $0 < w < 1$ represents the dominant Gaussian core, and $k > 1$ scales the variance to capture the heavier tails of the systematic offsets. The full likelihood of the data given the model parameters $\theta$ becomes a mixture of Gaussians with the corresponding standard deviations in bin $i$ defined as: $\sigma_{1, i} = \sqrt{dF_{\text{stat}, i}^2 + dF_{\text{syst}, i}^2}$ and $\sigma_{2, i} = \sqrt{dF_{\text{stat}, i}^2 + k^2 dF_{\text{syst}, i}^2}$. Thus the updated log-likelihood simplifies to: $\ln \mathcal{L} = -\frac{\chi^2(\sigma_1)}{2} + \ln \left[ 1 + \frac{1-w}{w} \exp\left( \frac{\chi^2(\sigma_1) - \chi^2(\sigma_2)}{2} \right) \right]$. The purely Gaussian model is recovered by setting $w = 1$ (with arbitrary $k > 0$) or setting $k = 1$ (with arbitrary $0 < w < 1$). It can be seen that a larger $k$ and smaller $w$ increases the width of the posterior marginal distribution, especially for the well-constrained parameters ($k^{0}_{\parallel}$, $a_{\perp}$, and $b_{\perp}$).}{Note that the likelihood remains the same as the baseline Gaussian noise model for $w = 1$ (with arbitrary $k > 0$) and $k = 1$ (with arbitrary $0 < w < 1$). These cases correspond to cells in the top row and leftmost column of each plot, respectively. Since in these cells the width is the same as in the ($w=1$, $k=1$) cell, we left them blank.}}
   \label{fig:noise_model_ams02}
\end{figure}

\add{
In our original formulation, we assumed that both the statistical uncertainties ($dF_{\text{stat}}$) and the systematic uncertainties ($dF_{\text{syst}}$) were normally distributed and uncorrelated, and thus the standard deviation for a given bin $r$ was defined as $\sigma_r = \sqrt{dF_{\text{stat}, r}^2 + dF_{\text{syst}, r}^2}$, leading to a standard $\chi^2$ log-likelihood (Section \ref{sec:description_of_problem}).
We now treat the measured flux $F_r$ as a random variable distributed as the convolution of a Normal distribution (representing statistical noise) and a fat-tailed distribution $S$ (representing systematic noise).
$S$ is modeled as a Gaussian Mixture Model (GMM) composed of a weighted sum of two Normal distributions with the same mean but different standard deviations: $S(\delta) = w \cdot \mathcal{N}(\delta; 0, dF_{\text{syst}}) + (1-w) \cdot \mathcal{N}(\delta; 0, k \cdot dF_{\text{syst}})$ where $0 < w < 1$ represents the dominant Gaussian core, and $k > 1$ scales the variance to capture the heavier tails of the systematic offsets.
The full likelihood of the data given the model parameters $\theta$ becomes a mixture of Gaussians with the corresponding standard deviations in bin $r$ defined as: $\sigma_{1, r} = \sqrt{dF_{\text{stat}, r}^2 + dF_{\text{syst}, r}^2}$ and $\sigma_{2, r} = \sqrt{dF_{\text{stat}, r}^2 + k^2 dF_{\text{syst}, r}^2}$.
Thus, after dropping normalization constants, the updated log-likelihood simplifies to:
\protect \begin{equation}
\ln \mathcal{L} = -\frac{\chi^2(\sigma_1)}{2} + \ln \left[ 1 + \frac{1-w}{w} \exp\left( \frac{\chi^2(\sigma_1) - \chi^2(\sigma_2)}{2} \right) \right],
\end{equation}}
\noindent
\add{where $\chi^2(\sigma) = \sum_r \,(x_r - S(\theta)_r)^2 / \sigma_r^2$.
The purely Gaussian model is recovered by setting $w = 1$ (with arbitrary $k > 0$) or setting $k = 1$ (with arbitrary $0 < w < 1$).
Figure~\ref{fig:noise_model_ams02} shows how the marginal posterior distribution width (standard deviation / mean) increases for larger $k$ and smaller $w$, especially for the well-constrained parameters ($k^{0}_{\parallel}$, $a_{\perp}$, and $b_{\perp}$). When $w$ and $k$ are unknown, they can be treated as nuisance parameters.
}

%
%

\section*{Open Research Section}
\label{sec:open-research-section}

The dataset and posterior samples used for constraining the transport parameters of GCRs in this study are available on GitHub (\url{https://github.com/hawaii-ai/GalacticCosmicRays}) via Zenodo with a Creative Commons Attribution license and public access (\url{https://doi.org/10.5281/zenodo.18249366}; \citeA{wolniewicz2026dataset}).


\section*{Conflict of Interest declaration}
The authors declare there are no conflicts of interest for this manuscript.

\acknowledgments
This material is based upon work supported by the National Science Foundation Graduate Research Fellowship Program under Grant 2236415, NSF 2238375, and the National Aeronautics and Space Administration under Grant Living with the Star No. LWS80NSSC20K1819. Any opinions, findings, conclusions, or recommendations expressed in this material are those of the author(s) and do not necessarily reflect the views of the National Science Foundation or the National Aeronautics and Space Administration. We acknowledge use of NASA/GSFC's Space Physics Data Facility's OMNIWeb service, and OMNI data. The authors would also like to thank Arianna Bunnell, Yannik Glaser, Yusuke Hatanaka, and Hans Mertens for their helpful conversations while preparing this manuscript. The technical support and advanced computing resources from University of Hawaii Information Technology Services, funded in part by the National Science Foundation CC* awards 2201428 and 2232862 are gratefully acknowledged. Resources supporting this work were provided by the NASA High-End Computing (HEC) Program through the NASA Advanced Supercomputing (NAS) Division at Ames Research Center.


\bibliography{references}

\end{document}